\documentclass[runningheads, orivec]{llncs}
\usepackage{amsmath}
\usepackage[T1]{fontenc}
\usepackage[whole]{bxcjkjatype}
\usepackage[most]{tcolorbox}
\usepackage{array}
\usepackage{makecell}
\usepackage{bbding}

\newtcolorbox{promptbox}{
  colback=gray!5,
  colframe=gray!60,
  boxrule=0.5pt,
  arc=2pt,
  left=4pt,
  right=4pt,
  top=4pt,
  bottom=4pt,
  fontupper=\small,
  breakable
}

\usepackage{graphicx}

\begin{document}

\title{Evaluating the Robustness of Japanese LLMs to IME-Related and Typographical Errors}

\author{
Ryota~Mibayashi\inst{1}\Envelope\orcidID{0009-0005-8995-5579} \and
Hiroaki~Ohshima\inst{2}\orcidID{0000-0002-9492-2246}
}
\authorrunning{R. Mibayashi and H. Ohshima}
\titlerunning{Robustness of Japanese LLMs to Typographical Errors}
\institute{
Kobe University, Hyogo, Japan
\email{mibayashi@people.kobe-u.ac.jp}
\and
University of Hyogo, Hyogo, Japan
\email{ohshima@ai.u-hyogo.ac.jp}
}

\maketitle              

\begin{abstract}
Large language models (LLMs) have achieved strong performance across various natural language processing tasks.
However, their robustness to typographical errors remains underexplored, particularly in Japanese, where text input involves multiple writing systems and IME-based conversion.
In this study, we evaluate the robustness of Japanese LLMs against realistic Japanese-specific typos.
We introduce five typo categories: Character Transposition, Character Replacement, Homophone Conversion, Japanese IME Conversion, and Full-Width Conversion. These perturbations are applied to three Japanese benchmark datasets (JMMLU, JCommonsenseQA, and JamC-QA), and eleven Japanese and multilingual LLMs are evaluated.
The results show that Character Transposition and Character Replacement typos consistently reduce accuracy across benchmarks, whereas IME Conversion, Full-Width Conversion, and Homophone Conversion have relatively limited impact. 
These findings reveal that current Japanese LLMs remain vulnerable to realistic Japanese typing errors, particularly those that substantially distort the original input, highlighting the importance of robustness evaluation in practical input environments.

\end{abstract}

\section{Introduction}

\begin{figure}
\includegraphics[width=\textwidth]{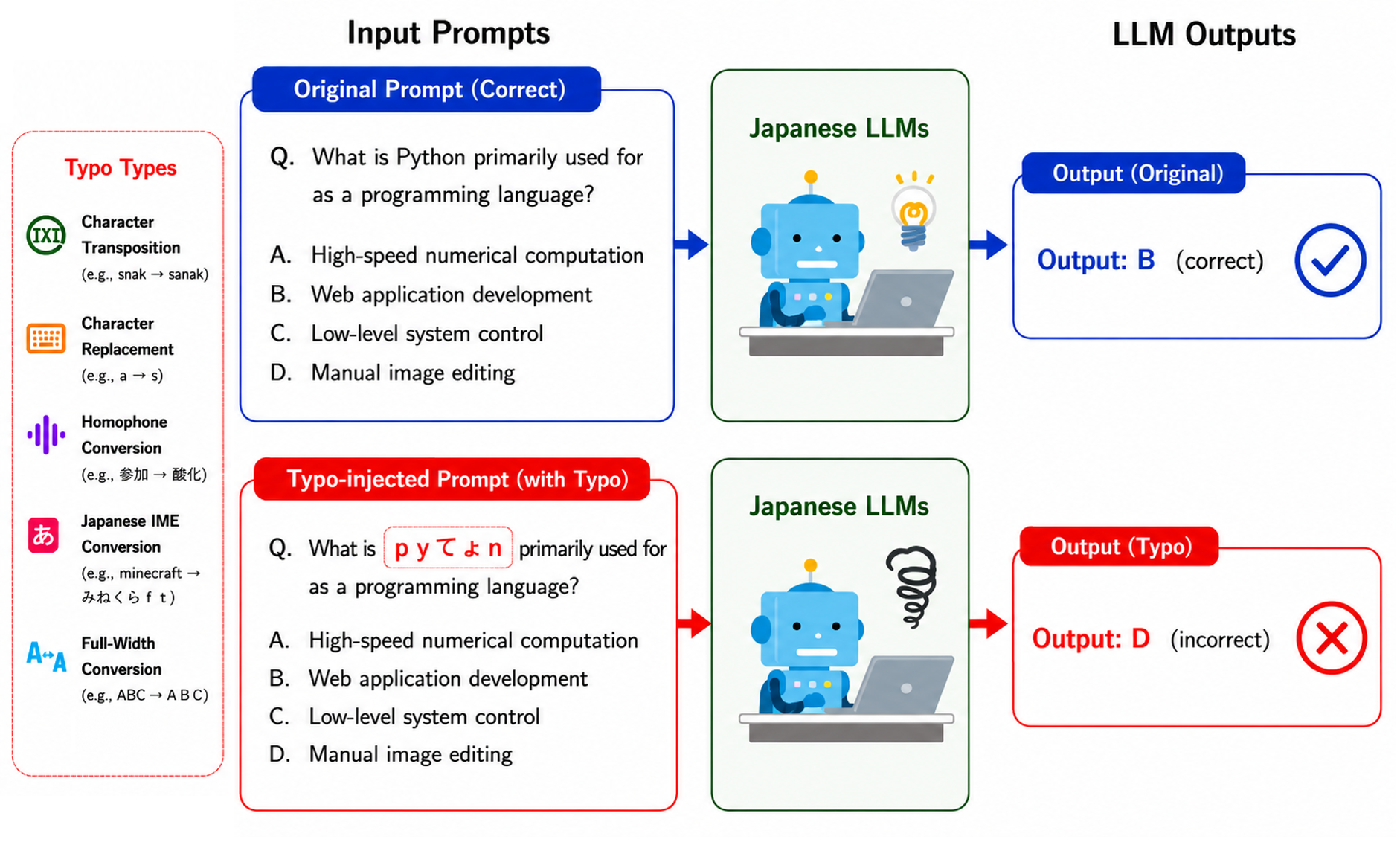}
\caption{Overview of the proposed evaluation framework. Original and typo-injected prompts are input to Japanese LLMs, and the resulting outputs are compared to measure the impact of different typo patterns on model performance.} \label{fig:overview}
\end{figure}

Large Language Models (LLMs) have achieved remarkable performance across a wide range of natural language processing tasks and are increasingly being deployed in real-world applications.
However, real-world user inputs frequently contain typographical errors and input mistakes.
Therefore, evaluating the robustness of LLMs against noisy inputs has become an important research problem.
Previous studies have investigated the impact of intentionally injected typographical errors in prompts on LLM performance~\cite{zhu2024promptrobust}.
These studies mainly focus on English and employ character-level attacks such as character insertion, deletion, replacement, and transposition.
Previous studies have shown that such character-level perturbations can not only degrade task performance but also alter model behavior, potentially enabling the circumvention of safety mechanisms such as response control and content moderation.
Therefore, evaluating the robustness of LLMs against typographical errors is important not only for maintaining task performance but also for ensuring the reliability and safety of deployed systems.

However, typographical errors in Japanese differ substantially from those in English.
While English typographical errors are typically represented as local modifications to alphabetic character sequences, Japanese text consists of multiple writing systems, including kanji, hiragana, katakana, and Roman characters.

Furthermore, Japanese text input commonly relies on an Input Method Editor (IME) for kana-to-kanji conversion, resulting in a wider variety of error sources.
For example, commonly observed Japanese input errors include IME conversion mistakes, in which the same phonetic input is converted into a different kanji expression with an unintended meaning. For example, the input ``\underline{kikaigakushuu}'' may be correctly converted to ``\underline{機械学習}'' (machine learning) but mistakenly converted to ``\underline{機会学習}'' (learning opportunities).
Another characteristic example arises from keyboard-position errors. The intended Romanized input ``\underline{sanka}'' (participation; ``\underline{参加}'') may become ``\underline{sanak}'' due to a single mistyped character. While this corresponds to only a one-character error in the input sequence, the subsequent IME conversion can produce a drastically different output such as ``\underline{三亜k}''. Unlike English typographical errors, which are often limited to local character-level modifications, Japanese typing errors can propagate through the IME conversion process and result in substantial semantic changes.
Mixed-script inputs that combine Roman and kana characters, such as writing ``\underline{python}'' as ``\underline{ｐｙてょｎ},'' can also occur.
These errors are not merely character-level perturbations but rather Japanese-specific noise arising from phonological and orthographic characteristics.
Therefore, evaluating the robustness of Japanese LLMs requires considering not only conventional character-level attacks developed for English but also Japanese-specific orthographic variations and phonological errors.

To address this issue, we propose Japanese-specific typo generation methods that simulate realistic input errors observed in Japanese text entry. As shown in Figure~\ref{fig:overview}, the generated typos are applied to existing Japanese benchmarks to evaluate their impact on the performance of Japanese LLMs.
Specifically, we define five typo patterns: Character Transposition, Character Replacement, Homophone Conversion, Japanese IME Conversion, and Full-Width Conversion, and apply them to existing Japanese benchmarks.
Using these perturbations, we quantitatively analyze which types of typographical errors Japanese LLMs are robust to and which types they are vulnerable to.
The contributions of this study are summarized as follows.

\begin{itemize}
\item We propose typo generation methods specifically designed for Japanese, taking into account realistic Japanese input environments.
\item We evaluate multiple Japanese LLMs and reveal their robustness to typographical errors by analyzing performance degradation across different typo categories.
\end{itemize}

\section{Related Work}
\subsection{Large Language Models}
Recent LLMs are typically based on the Transformer architecture~\cite{vaswani2017transformer}, with the GPT family~\cite{brown2020language,openai2024gpt4} and the Llama family~\cite{touvron2023llama,touvron2023llama2} being among the most widely used model series.
Recently, several models have been developed to improve Japanese language capabilities by continually pretraining and instruction-tuning these foundation models on Japanese corpora.
Representative examples include Swallow and ELYZA, which are based on the Llama architecture, as well as Sarashina and llm-jp, which are trained on large-scale Japanese corpora.
In addition, open-weight models such as GPT-OSS have recently become available, enabling comparisons across different architectures and training strategies.
In this study, we evaluate the typo robustness of multiple LLMs, including Ministral, Llama-based models, GPT-OSS-based models, and Japanese-specific models, in order to investigate the impact of model architecture and training strategies on robustness to Japanese input errors.

\subsection{Evaluation Benchmarks for LLMs}
Various benchmarks have been proposed to evaluate the performance of LLMs.
Representative examples include MMLU~\cite{hendrycks2021measuring} for evaluating multidisciplinary knowledge and BIG-bench~\cite{srivastava2023beyond} for assessing reasoning abilities. For Japanese LLMs, benchmarks such as JMMLU~\cite{yin2024respect}, JCommonsenseQA~\cite{kurihara2022jglue}, and JamC-QA~\cite{oka2026jamcqa} have been proposed to evaluate knowledge, commonsense reasoning, and Japanese cultural understanding.

Several datasets have also been developed for Japanese typographical errors.
For example, JWTD (Japanese Wikipedia Typo Dataset)~\cite{tanaka2020building} was constructed from Wikipedia revision histories and provides typo–correction pairs for evaluating Japanese typo correction systems.
In addition, Hagiwara et al.~\cite{Hagiwara2019githubtypo} constructed a multilingual typo corpus from GitHub revision histories, providing large-scale datasets of typographical errors and their corrected forms for multiple languages, including Japanese.
However, existing Japanese typo datasets are primarily intended for typo correction and text normalization.
As a result, they are not suitable for evaluating how typographical errors affect downstream task performance or the robustness of LLMs.

Therefore, instead of evaluating typo correction itself, we inject realistic Japanese typo patterns into existing multiple-choice benchmarks with ground-truth answers, namely JMMLU, JCommonsenseQA, and JamC-QA, and evaluate the robustness of Japanese LLMs against such input noise.

\subsection{Robustness to Typographical Errors and Noisy Inputs}
Traditional spelling correction studies have commonly modeled misspellings using character-level edit operations, including insertion, deletion, substitution, and transposition~\cite{damerau1964technique,kukich1992techniques}.
More recently, researchers have actively investigated the robustness of large language models to typographical errors.
Kumar et al.~\cite{kumar2020noisy} reported that natural misspellings can significantly degrade the performance of language models.
Furthermore, Gan et al.~\cite{gan2024reasoning} demonstrated that injecting only a small number of typographical errors into reasoning tasks can substantially reduce the reasoning performance of LLMs.
PromptRobust~\cite{zhu2024promptrobust} systematically evaluated the impact of prompt perturbations, including typographical errors and paraphrasing, on LLM performance.
However, most of these studies focus primarily on English and pay limited attention to Japanese-specific input errors.

Furthermore, studies have also been conducted for Chinese, focusing on input-method-related noise.
READIN~\cite{si2023readin} is a benchmark for realistic input noise in Chinese, including pinyin input mistakes and speech-recognition errors.
While our work shares the same motivation of evaluating input-environment-induced noise, it differs from READIN by focusing on Japanese-specific typo patterns.

\subsection{Adversarial Typo Attacks}
A large body of research has investigated adversarial attacks to analyze the vulnerabilities of LLMs and other natural language processing models.
TextBugger~\cite{li2019textbugger} and DeepWordBug~\cite{gao2018black} proposed methods that mislead model predictions through small character-level perturbations, such as insertion, deletion, substitution, and transposition.
TextFooler~\cite{jin2020bert} and BERT-Attack~\cite{li2020bertattack} introduced word-level attacks based on synonym substitution.

However, most of these attack methods were designed for alphabet-based languages, and their applicability to languages with multiple writing systems, such as Japanese, has not been sufficiently explored.

\subsection{Typo Taxonomy}
Several studies have proposed systematic categorizations of typographical errors and orthographic variations.
Sperduti et al.~\cite{sperduti2025misspellings} presented a comprehensive survey of misspellings in NLP, covering misspelling generation, spelling correction, adversarial misspellings, robustness to noisy text, and evaluation benchmarks.

However, most existing taxonomies were developed for English and do not adequately consider the unique characteristics of Japanese input environments.
Japanese contains unique typo patterns arising from its input methods and writing systems, including IME conversion errors, Roman-input-derived mistakes, omission of voicing marks, confusion of small kana characters, and omission of long vowels.
Therefore, this study introduces a typo taxonomy tailored to Japanese input environments and evaluates the robustness of Japanese LLMs based on this taxonomy.

\section{Japanese-Specific Typo Injection}
To evaluate the typo robustness of Japanese LLMs, we propose a typo injection method that simulates realistic input errors occurring in Japanese text entry environments.
Previous typo studies have primarily focused on English and have widely employed character-level attacks, such as character insertion, deletion, substitution, and transposition, as exemplified by TextBugger~\cite{li2019textbugger} and DeepWordBug~\cite{gao2018black}.
However, these methods were primarily designed for alphabet-based languages and do not adequately reflect the characteristics of Japanese text input environments.
Therefore, we design typo patterns specific to Japanese text input based on keyboard-based input environments.
The typo categories used in this study are summarized in Table~\ref{tab:typo_examples} and described below.
\begin{itemize}
    \item Character Transposition Typo
    \item Character Replacement Typo
    \item Homophone Conversion Typo
    \item Japanese IME Conversion Typo
    \item Full-Width Conversion Typo
\end{itemize}

The details of these typo categories and their transformation methods are presented in the next subsection.
The specific prompts used in our experiments are provided in Appendix A.

\subsubsection{Character Transposition Typo}
This typo simulates cases in which the order of keystrokes is mistakenly reversed during Roman-character input.
For example, as shown in Table~\ref{tab:typo_examples}, when typing ``san\underline{ka}'', the user may mistakenly enter ``san\underline{ak}'' by reversing the order of the characters ``\underline{a}'' and ``\underline{k}''.
As a result, instead of the intended conversion ``sanka → 参加(participation)'' , the mistyped input is converted as ``sanak → 三亜k (nonsensical output)'', producing an unnatural string.
This typo is intended to reproduce a common input error that occurs during Japanese keyboard-based text entry.

To generate this typo, the target Japanese text span in a sentence was first converted into its Romanized form using pykakasi.
Two key positions corresponding to vowels or consonants in the Romanized string were then randomly selected, and the characters at those positions were swapped.
The resulting Romanized string was subsequently converted back into kana and further transformed into a mixed kanji–kana expression to simulate IME-based conversion.

\begin{table}[tb]
\centering
\caption{Examples of the proposed typo transformations}
\label{tab:typo_examples}
\begin{tabular}{l|l}
\hline\hline
Typo Type & Example \\
\hline

Character Transposition &
\makecell[l]{
\textbf{Original:} san\underline{ka} $\rightarrow$ 参加 (participation) \\
\textbf{Typo:\hspace{1.25em}} san\underline{ak} $\rightarrow$ 三亜k (nonsensical output)
} \\\hline

Character Replacement &
\makecell[l]{
\textbf{Original:} nose\underline{r}u $\rightarrow$ 乗せる (place on) \\
\textbf{Typo:\hspace{1.25em}} nose\underline{t}u $\rightarrow$ の節 (phrase)
} \\\hline

Homophone Conversion&
\makecell[l]{
\textbf{Original:} sanka $\rightarrow$ 参加 (participation) \\
\textbf{Typo:\hspace{1.25em}} sanka $\rightarrow$ 酸化 (oxidation)
} \\\hline

Japanese IME Conversion &
\makecell[l]{
\textbf{Original:} minecraft $\rightarrow$ minecraft \\
\textbf{Typo:\hspace{1.25em}} minecraft $\rightarrow$ みねｃらｆｔ (nonsensical output)
} \\\hline

Full-Width Conversion &
\makecell[l]{
\textbf{Original:} AI $\rightarrow$ AI \\
\textbf{Typo:\hspace{1.25em}} AI $\rightarrow$ ＡＩ
} \\\hline

\hline\hline
\end{tabular}
\end{table}

\vspace{-0.5em}

\subsubsection{Character Replacement Typo}
This typo simulates cases in which an incorrect key is pressed during Roman-character input.
For example, a user may mistakenly type ``nose\underline{t}u'' instead of the intended input ``nose\underline{r}u.''
As a result, instead of the intended conversion ``noseru'' → ``乗せる (place on)'', the mistyped input is converted as ``nosetu'' → ``の節 (phrase)'', producing an unnatural string.
Similar to the previously described transposition typo, this typo is designed to simulate a common input mistake in Japanese keyboard-based text entry.

To generate this typo, the target Japanese text span in a sentence was first converted into its Romanized form using pykakasi.
A key position corresponding to either a vowel or a consonant in the Romanized string was then randomly selected, and the character at that position was replaced with another character of the same type.
The modified Romanized string was subsequently converted back into kana and further transformed into a mixed kanji–kana expression to simulate IME-based conversion.

\subsubsection{Homophone Conversion Typo}
This typo simulates conversion errors in which an incorrect candidate is selected among words sharing the same pronunciation.
For example, the input ``sanka'' may be mistakenly converted to ``酸化(sanka/oxidation)'' instead of the intended ``参加(sanka/participation)''.
Because this transformation preserves pronunciation while altering lexical meaning, it allows us to examine how semantic changes introduced by IME conversion affect model performance.

To generate this typo, we simulated incorrect kanji conversions in which a word is replaced by another word with the same pronunciation.
We predefined sets of homophonous words, such as ``回答／解答'', ``対象／対照'', ``機械／器械／機会'', ``確率／格率'', and ``分布／分賦''.
When such words appeared in the text, 10\% of the candidate instances were randomly selected and replaced with another word from the same homophone set.
This setting generates inputs containing conversion errors that differ orthographically while remaining phonologically identical or highly similar.

\vspace{-0.5em}

\subsubsection{Japanese IME Conversion Typo}
This typo simulates situations in which users accidentally type English words while remaining in Japanese IME mode.
For example, when typing ``minecraft,'' an active IME may produce a mixed-script expression such as ``みねｃらｆｔ.''
This is a common input error frequently encountered by Japanese users. Because IME conversion is inadvertently applied to strings that do not require conversion, the resulting text often obscures the original English word.

To generate this typo, we simulated situations in which romaji input and full-width English character input are mixed during Japanese text entry.
For each selected ASCII word, the prefix that could be greedily interpreted as romaji from the beginning was converted into kana, while the remaining alphanumeric characters were converted into full-width characters.

\vspace{-0.5em}

\subsubsection{Full-Width Conversion Typo}
This typo simulates orthographic variations caused by the use of full-width and half-width characters.
For example, the string ``AI'' may be converted into its full-width counterpart ``ＡＩ''.
Although this transformation causes little semantic change, it may affect tokenization and embedding representations, making it useful for evaluating robustness to orthographic variation.

To generate this typo, half-width alphanumeric characters were converted into their full-width counterparts.
For example, ``BERT'' and ``2024'' were converted into ``ＢＥＲＴ'' and ``２０２４'', respectively.
This transformation does not modify the Japanese text itself and introduces only variations in alphanumeric representation.

\section{Experiments}

\subsection{Evaluation Datasets}
In this study, we use multiple Japanese benchmark datasets covering knowledge understanding, commonsense reasoning, and Japanese cultural knowledge to evaluate the robustness of Japanese LLMs to typographical errors.
While typographical robustness is commonly evaluated using MMLU in English, the effects of typographical errors are expected to be more complex in Japanese due to its diverse writing systems, including kanji, kana, romaji, long vowels, and voiced sound marks.
Therefore, we conduct our evaluation using a diverse set of Japanese benchmark datasets, namely JMMLU~\cite{yin2024respect}, JCommonsenseQA~\cite{kurihara2022jglue}, and JamC-QA~\cite{oka2026jamcqa}~\footnote{~\url{https://huggingface.co/datasets/sbintuitions/JamC-QA}}.
All of these benchmarks are four-choice multiple-choice tasks, allowing unified evaluation by extracting the final selected option from the model outputs.

In this study, we randomly sampled 128 examples from each of the three benchmarks, resulting in a total of 384 samples. We then applied five types of typographical transformations to each sample.
Typographical perturbations were applied only to the question text, while the prompt template and answer choices remained unchanged.
In addition, typo perturbations were applied to 10\% of the eligible characters or words in each input, with at least one typo introduced per sample.

\subsection{Evaluation  Models}
In this study, we evaluate the robustness of LLMs to typographical errors.
To this end, we use both Japanese-specialized LLMs and multilingual LLMs with model sizes ranging from 0.5B to 20B parameters.
The evaluated models are as follows.

\begin{itemize}
    \item Ministral-8B-Instruct-2410
    \item Llama-3.1-Swallow-8B-Instruct-v0.5
    \item Llama-3-ELYZA-JP-8B
    \item llm-jp-3-13b-instruct3
    \item llm-jp-3-7.2b-instruct3
    \item Sarashina2.2-3b-instruct-v0.1
    \item Sarashina2.2-1b-instruct-v0.1
    \item Sarashina2.2-0.5b-instruct-v0.1
    \item GPT-OSS-20B
    \item GPT-OSS-Swallow-20B-SFT-v0.1
    \item GPT-OSS-Swallow-20B-RL-v0.1
\end{itemize}

These models have different parameter scales ranging from approximately 0.5B to 20B, enabling an analysis of the impact of model size on robustness to Japanese typographical errors.
In addition, the Sarashina, Swallow, ELYZA, and llm-jp series are Japanese-specialized LLMs developed through continued pre-training on Japanese corpora and/or Japanese instruction tuning, with the aim of improving Japanese understanding and generation capabilities.
Furthermore, GPT-OSS-Swallow-RL is a model that has undergone additional training using reinforcement learning and is expected to achieve improved reasoning ability and Japanese conversational performance.
Moreover, these models differ in several design aspects, including Japanese adaptation strategies, tokenizer design, training corpora, and model size.
Therefore, this study compares and analyzes how differences in these design factors affect robustness to Japanese typographical errors.

In addition, regarding the maximum generation length, non-reasoning models were evaluated with a generation length of 64 tokens due to computational resource constraints.
For reasoning models (GPT-OSS-20B, GPT-OSS-Swallow-20B-SFT-v0.1, and GPT-OSS-Swallow-20B-RL-v0.1), the generation length was set to 1024 tokens because their intermediate reasoning processes can become lengthy, causing them to fail to produce a final answer within shorter generation limits.

\subsection{Evaluation Protocol}
Accuracy was used as the evaluation metric.
The answer extraction procedure for model outputs was defined as follows.
We first extracted the option A/B/C/D appearing after the Japanese keywords corresponding to ``answer'' (e.g., ``回答'', ``答え'', or ``正解'').
If multiple candidates were found, the last occurring option was selected.
If no such pattern was found, we searched for a standalone A/B/C/D at the end of the generated text.
If extraction still failed, the first standalone occurrence of A/B/C/D in the generated text was adopted as the model's answer.

\subsection{Prompt Construction}
A common prompt format was used for all tasks, as shown below.
In addition, the prompts were constructed using five-shot examples.
These five-shot examples were taken from the original, typo-free dataset, and examples identical to the target question were excluded.
Although the example below is presented in English for explanatory purposes, the actual prompts used in the experiments were written in Japanese.

{\small
\begin{promptbox}
\textbf{Prompt:}\\
Please answer the following question.

Output only a single character: A, B, C, or D.

Do not output any additional text.

The following are example answers.

Question:

Who fits in the blank in the following sentence?

"In 1252, \{ \} was appointed as the sixth shogun."

Choices:

A. Prince Mochihito, B. Prince Munetaka, C. Saionji Kintsune, D. Kujo Yoritsugu

Answer: B

---

Question:

A change in the membrane potential of a neuron from +50 mV to -70 mV is considered what?

Choices:

A. Depolarization, B. Repolarization, C. Hyperpolarization, D. Action potential

Answer: B

---

Question:

How many numbers are contained in the list $25, 26, 27, \ldots, 99, 100$?

Choices:

A. 75, B. 76, C. 22, D. 23

Answer: B

---

Question:

Which of the following best describes the fallacy of style over substance?

Choices:

A. Changing the meaning of a word or phrase from one part of an argument to another.

B. Confusing figurative language with literal language.

C. Inferring that an idea is correct because it is presented in an attractive manner, regardless of its content.

D. Using emotionally charged language to create an impression about a claim without discussing whether the language is relevant to the subject.

Answer: C

---

Question:

According to Cutlip (2006), at which stage of communication planning should one determine what to say and why it should be said?

Choices:

A. Situation Analysis

B. Strategy

C. Implementation

D. Evaluation, in which all aspects of the program are assessed

Answer: B

---

Now answer the following question.

Question:

Statement 1: Unlike logistic regression models, support vector machines do not provide a probability distribution over possible labels for a given input instance.

Statement 2: When moving from a linear kernel to a higher-order polynomial kernel, the support vectors are generally expected to remain unchanged.

Choices:

A. True, True, B. False, False, C. True, False, D. False, True

Answer:

\end{promptbox}
}

\section{Results and Discussion}

\begin{table}[tb]
\centering
\caption{
Accuracy of each model under different Japanese typo conditions.
Orig. denotes the original prompts without typos.
Values lower than the original accuracy are underlined.
Average represents the mean accuracy across all evaluated models.
}
\label{tab:all_typo}
\small
\begin{tabular}{p{0.20\linewidth}|*{6}{>{\centering\arraybackslash}p{0.12\linewidth}}}

\hline
Model & Orig. & Trans. & Repl. & Homo. & IME & FullW. \\
\hline

Ministral-8B
& 0.583
& \underline{0.565}
& \underline{0.503}
& \underline{0.578}
& \underline{0.576}
& \underline{0.573} \\

Swallow-8B
& 0.677
& \underline{0.607}
& \underline{0.615}
& \underline{0.672}
& 0.677
& \underline{0.672} \\

ELYZA-8B
& 0.568
& \underline{0.529}
& \underline{0.516}
& \underline{0.560}
& \underline{0.565}
& \underline{0.565} \\

llm-jp-13B
& 0.615
& \underline{0.578}
& \underline{0.531}
& \underline{0.602}
& \underline{0.607}
& \underline{0.609} \\

llm-jp-7.2B
& 0.578
& \underline{0.513}
& \underline{0.477}
& 0.578
& 0.578
& 0.581 \\

Sarashina-3B
& 0.698
& \underline{0.620}
& \underline{0.638}
& \underline{0.688}
& \underline{0.695}
& 0.701 \\

Sarashina-1B
& 0.438
& \underline{0.419}
& \underline{0.406}
& \underline{0.435}
& \underline{0.435}
& 0.440 \\

Sarashina-0.5B
& 0.352
& \underline{0.333}
& \underline{0.331}
& 0.354
& \underline{0.346}
& 0.354 \\

GPT-OSS-20B
& 0.708
& \underline{0.643}
& \underline{0.615}
& 0.711
& \underline{0.701}
& \underline{0.703} \\

GPT-OSS-SFT
& 0.784
& \underline{0.732}
& \underline{0.721}
& \underline{0.771}
& 0.786
& \underline{0.771} \\

GPT-OSS-RL
& 0.703
& \underline{0.641}
& \underline{0.607}
& 0.708
& 0.732
& 0.740 \\

\hline
Average
& 0.609
& \underline{0.562}
& \underline{0.542}
& \underline{0.605}
& 0.609
& 0.610 \\
\hline

\end{tabular}
\end{table}

\begin{figure}[tb]
\includegraphics[width=\linewidth]{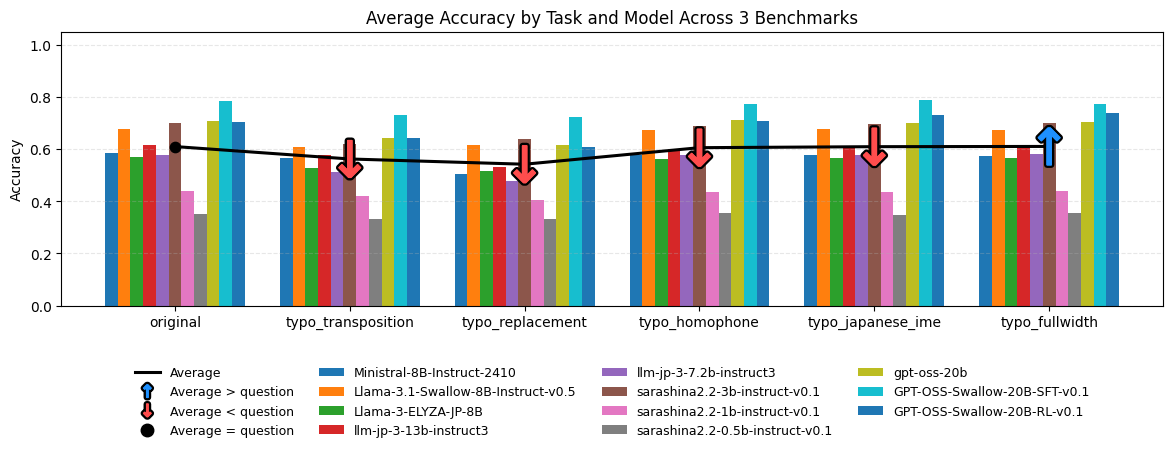}
\caption{
Accuracy of each model under different typo conditions averaged over the three benchmarks.
The black line indicates the average accuracy across all evaluated models.
Arrows indicate changes relative to the original prompts.
}
\label{fig:result}
\end{figure}

Figure~\ref{fig:result} and Table~\ref{tab:all_typo} show the average results across the three benchmarks.
Overall, the average accuracy under the Original condition was 0.609, whereas all typo conditions except Full-Width Conversion resulted in lower accuracy.
These results indicate that even Japanese-specialized LLMs are not robust to the typo conditions proposed in this study, resulting in decreased performance.

Notably, the Transposition and Replacement conditions resulted in the largest performance declines, with accuracy drops of 0.047 and 0.067, respectively, relative to the Original condition.
Specifically, under the Transposition condition, a question such as ``SVMに\underline{関し (kansi)}て正しいのはどれか？/Which of the following statements \underline{about} SVM is correct?'' could be transformed into ``SVMに\underline{k名氏 (knasi)}て正しいのはどれか？'', where the Romanized input ``kansi'' was mistyped as ``knasi''. As a result, the phrase ``関し/about'' was converted into the meaningless string ``k名氏/nonsensical'', making the question difficult to interpret correctly.
However, it may still be possible to infer the correct answer from contextual cues such as ``SVM'' and the phrase ``Which of the following statements is correct?''. Therefore, the robustness of LLMs to such typographical errors remains an open issue.

Specifically, under the Replacement condition, a question such as ``略語ではない\underline{単語 (tango)}を選べ/Select the \underline{word} that is not an abbreviation'' could be transformed into ``略語ではない\underline{派n後 (hango)}を選べ'', where the Romanized input ``tango'' was mistyped as ``hango''.
As a result, the phrase ``単語/word'' was converted into the meaningless string ``派n後/nonsensical'', making the question difficult to interpret correctly.
In the Replacement condition, the accuracy decreased substantially because the typo introduced characters that differed from the intended input, making it more difficult for the LLM to correctly understand the original meaning of the text.

Next, the accuracy drop under the Homophone condition was only 0.004 on average, and some models exhibited no degradation at all.
Interestingly, although homophone substitution appears to be a typo that substantially alters the meaning of a sentence, its impact on task accuracy was minimal.
Specifically, a question such as ``2024年\underline{現在}、任天堂のゲーム機で8ボタンのコントローラーはどれ？/As of 2024, which Nintendo game console has an eight-button controller?'' could be transformed into ``2024年\underline{原罪}、任天堂のゲーム機で8ボタンのコントローラーはどれ？'', where the word ``現在/currently (\textit{genzai})'' was incorrectly converted to ``原罪/original sin'' (\textit{genzai})''. 
Despite this substantial semantic change, the LLM was still able to answer the question correctly.
Given the negligible impact on accuracy, this result suggests that many questions remained answerable despite homophone substitutions, possibly because sufficient information was preserved in the surrounding context.

In contrast, the Japanese IME condition achieved the same accuracy as the Original condition.
Although this typo type often produced highly destructive transformations that failed to preserve the original form of the input, such as converting ``ImageNet'' into ``いまげねｔ'' or ``tanh'' into ``たｎｈ'', the LLMs were still able to answer questions containing such transformations correctly.
However, because most Japanese-specialized LLMs still exhibited some degree of performance degradation under this condition, the results suggest that Japanese-specialized models are not particularly robust to this type of typo.

Finally, the Full-Width condition achieved a slightly higher accuracy than the Original condition, with an increase of only 0.001.
However, this difference is negligible and can be regarded as noise, especially given that some models exhibited a decrease in performance under this condition.
In practice, many Full-Width transformations remained highly readable and were essentially variations in notation, such as converting ``ResNet'' to ``ＲｅｓＮｅｔ'' or ``2000年'' to ``２０００年''.
This suggests that such Full-Width variations are largely treated as orthographic variants rather than substantial semantic perturbations by LLMs.
The results averaged over the three benchmarks are shown in Table~\ref{tab:benchmark_typo} and Figure~\ref{fig:results} in Appendix B.

\section{Conclusion}
In this study, we evaluated the robustness of Japanese LLMs to realistic Japanese-specific typographical errors. We introduced five typo categories: Character Transposition, Character Replacement, Homophone Conversion, Japanese IME Conversion, and Full-Width Conversion. These perturbations were applied to three Japanese benchmark datasets, and eleven Japanese and multilingual LLMs were evaluated.
The experimental results showed that Character Transposition and Character Replacement consistently caused the largest performance degradation across models. These typo types often produced corrupted strings through the interaction between Romanized input errors and IME conversion, making it difficult for LLMs to recover the original meaning of the input.
In contrast, Homophone Conversion, Japanese IME Conversion, and Full-Width Conversion had only a limited impact on accuracy. 
In particular, the limited impact of Homophone Conversion indicates that many questions remained answerable even when individual words were replaced with semantically different homophones, possibly because sufficient contextual information was preserved elsewhere in the input. 
Overall, our findings indicate that current Japanese LLMs remain vulnerable to typo patterns that substantially distort the surface form of the input, while they are relatively robust to orthographic variations that preserve readability.
As future work, we plan to expand the evaluation to additional benchmarks and larger-scale datasets, as well as investigate how tokenization and internal representations contribute to robustness against Japanese typographical errors.

\begin{credits}

\subsubsection{\ackname} This work was supported by JSPS KAKENHI Grant Numbers JP25K03229, JP25K03228, and JP24K03228, and by ROIS NII Open Collaborative Research 2026-261S04-24208.

\subsubsection{\discintname}
The authors have no competing interests to declare that are relevant to the content of this article.

\subsubsection{Generative AI Usage Disclosure}
The authors used OpenAI ChatGPT 5.5 during the preparation of this manuscript to improve English phrasing and readability, and to support the creation of figures used for illustration. All outputs, including generated text and figures, were reviewed and, where appropriate, revised by the authors, who take full responsibility for the content of the manuscript.
\end{credits}

\clearpage
\appendix

\section{Typo Examples}
\label{app:prompts}

\begin{promptbox}
\textbf{Character Transposition Typo}\\

\textbf{ORIGINAL:}\\
以下のベイジアンネットワークを考えてみよう。このベイジアンネットワークH -> U <- P <- Wに必要な\textbf{【独立】}パラメータはいくつあるか?\\
\textit{(Consider the following Bayesian network. How many \textbf{【independent】} parameters are required for the Bayesian network H -> U <- P <- W?)}\\

\textbf{TYPO:}\\
以下のベイジアンネットワークを考えてみよう。このベイジアンネットワークH -> U <- P <- Wに必要な\textbf{【度区利tうs】}パラメータはいくつあるか?\\
\textit{(Consider the following Bayesian network. How many parameters are required for the Bayesian network H -> U <- P <- W? Here, \textbf{【度区利tうs】} is a corrupted form of \textbf{【independent】} .)}\\[1em]

\textbf{Character Replacement Typo}\\

\textbf{ORIGINAL:}\\
文1| RoBERTaは、BERTが事前学習したコーパスの約10倍大\textbf{【き】}いコーパスで事前学習する。文2| 2018 年の ResNeXts は\textbf{【通常】} tanh 活性化関数を使用した。\\
\textit{(Sentence 1| RoBERTa was pretrained on a corpus approximately ten times \textbf{【larger】} than that used for BERT. Sentence 2| ResNeXts in 2018 \textbf{【normally】} used the tanh activation function.)}\\

\textbf{TYPO:}\\
文1| RoBERTaは、BERTが事前学習したコーパスの約10倍大\textbf{【dい】}いコーパスで事前学習する。文2| 2018 年の ResNeXts は\textbf{【t部う時ょう】} tanh 活性化関数を使用した。\\
\textit{(Sentence 1| RoBERTa was pretrained on a corpus approximately ten times larger than that used for BERT, where \textbf{【dい】} is a corruption of part of the word \textbf{【larger】}. Sentence 2| ResNeXts in 2018 used the tanh activation function, where \textbf{【t部う時ょう】} is a corrupted form of \textbf{【normally】} .)}\\[1em]

\textbf{Homophone Conversion Typo}\\

\textbf{ORIGINAL:}\\
文1|サポート・ベクトル・マシンは、ロジスティック\textbf{【回帰】}モデルのように、入力例が与えられたときに可能なラベルの確率分布を与える。文2|線形カーネルから高次の多項式カーネルに移行しても、サポート・ベクトルは一般的に変わらないと予想される。\\
\textit{(Sentence 1| Support Vector Machines provide a probability distribution over possible labels, similar to logistic \textbf{【regression】} models. Sentence 2| It is generally expected that support vectors remain unchanged when moving from a linear kernel to a higher-order polynomial kernel.)}\\

\textbf{TYPO:}\\
文1|サポート・ベクトル・マシンは、ロジスティック\textbf{【会期】}モデルのように、入力例が与えられたときに可能なラベルの確率分布を与える。文2|線形カーネルから高次の多項式カーネルに移行しても、サポート・ベクトルは一般的に変わらないと予想される。\\
\textit{(Sentence 1| Support Vector Machines provide a probability distribution over possible labels, similar to logistic \textbf{【session period】} models. Sentence 2|  It is generally expected that support vectors remain unchanged when moving from a linear kernel to a higher-order polynomial kernel.)}\\[1em]

\textbf{Japanese IME Conversion Typo}\\

\textbf{ORIGINAL:}\\
文1| RoBERTaは、BERTが事前学習したコーパスの約10倍大きいコーパスで事前学習する。文2| 2018 年の ResNeXts は通常 \textbf{【tanh】} 活性化関数を使用した。\\
\textit{(Sentence 1| RoBERTa was pretrained on a corpus approximately ten times larger than that used for BERT. Sentence 2| ResNeXts in 2018 typically used the \textbf{【tanh】} activation function.)}\\

\textbf{TYPO:}\\
文1| RoBERTaは、BERTが事前学習したコーパスの約10倍大きいコーパスで事前学習する。文2| 2018 年の ResNeXts は通常 \textbf{【たｎｈ】} 活性化関数を使用した。\\
\textit{(Sentence 1| RoBERTa was pretrained on a corpus approximately ten times larger than that used for BERT. Sentence 2| ResNeXts in 2018 typically used the activation function \textbf{【たｎｈ】}, which contains an IME conversion error derived from \textbf{【tanh】}.)}\\[1em]

\textbf{Full-Width Conversion Typo}\\

\textbf{ORIGINAL:}\\
文1| RoBERTaは、BERTが事前学習したコーパスの約10倍大きいコーパスで事前学習する。文2| \textbf{【2018】} 年の ResNeXts は通常 tanh 活性化関数を使用した。\\
\textit{Sentence 1| RoBERTa was pretrained on a corpus approximately ten times larger than that used for BERT. Sentence 2| In \textbf{【2018】}, ResNeXts typically used the tanh activation function.}\\

\textbf{TYPO:}\\
文1| RoBERTaは、BERTが事前学習したコーパスの約10倍大きいコーパスで事前学習する。文2| \textbf{【２０１８】} 年の ResNeXts は通常 tanh 活性化関数を使用した。\\
\textit{Sentence 1| RoBERTa was pretrained on a corpus approximately ten times larger than that used for BERT. Sentence 2| In \textbf{【２０１８】}, ResNeXts typically used the tanh activation function, where the year is written using full-width characters.}
\end{promptbox}

\section{Benchmark-wise Results}
To examine whether the impact of different typo types varies across benchmarks, we further report the results separately for each benchmark.
Detailed results for each benchmark are presented in Table~\ref{tab:benchmark_typo}.

\begin{table}[ht]
\centering
\caption{Accuracy under different typo conditions for each benchmark.}
\label{tab:benchmark_typo}
\small
\setlength{\tabcolsep}{2pt}
\renewcommand{\arraystretch}{1.0}

\textbf{(a) JMMLU}

\resizebox{0.82\textwidth}{!}{
\begin{tabular}{p{0.20\linewidth}|*{6}{>{\centering\arraybackslash}p{0.12\linewidth}}}
\hline
Model & Orig. & Trans. & Repl. & Homo. & IME & FullW. \\
\hline
\hline

Ministral-8B
& 0.516
& \underline{0.508}
& \underline{0.406}
& \underline{0.508}
& \underline{0.508}
& \underline{0.500} \\

Swallow-8B
& 0.633
& \underline{0.586}
& \underline{0.602}
& \underline{0.625}
& \underline{0.625}
& \underline{0.609} \\

ELYZA-8B
& 0.461
& 0.477
& \underline{0.430}
& \underline{0.445}
& 0.461
& \underline{0.453} \\

llm-jp-13B
& 0.438
& \underline{0.430}
& \underline{0.398}
& 0.438
& \underline{0.422}
& \underline{0.422} \\

llm-jp-7.2B
& 0.406
& \underline{0.383}
& \underline{0.305}
& \underline{0.398}
& \underline{0.398}
& 0.406 \\

Sarashina-3B
& 0.570
& \underline{0.500}
& \underline{0.539}
& \underline{0.539}
& \underline{0.555}
& 0.570 \\

Sarashina-1B
& 0.328
& 0.336
& \underline{0.320}
& 0.336
& 0.328
& 0.328 \\

Sarashina-0.5B
& 0.313
& \underline{0.273}
& 0.313
& \underline{0.305}
& \underline{0.305}
& 0.313 \\

GPT-OSS-20B
& 0.883
& \underline{0.812}
& \underline{0.797}
& 0.891
& \underline{0.867}
& \underline{0.859} \\

GPT-OSS-SFT
& 0.859
& \underline{0.844}
& \underline{0.828}
& \underline{0.844}
& \underline{0.852}
& \underline{0.852} \\

GPT-OSS-RL
& 0.789
& \underline{0.773}
& 0.797
& 0.805
& 0.813
& 0.828 \\

\hline
Average
& 0.563
& \underline{0.538}
& \underline{0.494}
& \underline{0.558}
& \underline{0.557}
& \underline{0.558} \\
\hline
\end{tabular}
}

\vspace{1mm}

\textbf{(b) JCommonsenseQA}

\resizebox{0.82\textwidth}{!}{
\begin{tabular}{p{0.20\linewidth}|*{6}{>{\centering\arraybackslash}p{0.12\linewidth}}}
\hline
Model & Orig. & Trans. & Repl. & Homo. & IME & FullW. \\
\hline
\hline

Ministral-8B
& 0.883
& \underline{0.820}
& \underline{0.773}
& \underline{0.875}
& \underline{0.867}
& \underline{0.867} \\

Swallow-8B
& 0.938
& \underline{0.805}
& \underline{0.859}
& 0.938
& 0.938
& 0.938 \\

ELYZA-8B
& 0.844
& \underline{0.742}
& \underline{0.758}
& \underline{0.836}
& \underline{0.836}
& 0.844 \\

llm-jp-13B
& 0.898
& \underline{0.789}
& \underline{0.742}
& 0.898
& \underline{0.891}
& 0.898 \\

llm-jp-7.2B
& 0.820
& \underline{0.711}
& \underline{0.734}
& \underline{0.812}
& 0.828
& 0.828 \\

Sarashina-3B
& 0.945
& \underline{0.828}
& \underline{0.859}
& 0.945
& 0.945
& 0.945 \\

Sarashina-1B
& 0.672
& \underline{0.578}
& \underline{0.586}
& \underline{0.664}
& \underline{0.664}
& 0.672 \\

Sarashina-0.5B
& 0.469
& \underline{0.438}
& \underline{0.398}
& 0.469
& \underline{0.461}
& \underline{0.461} \\

GPT-OSS-20B
& 0.945
& \underline{0.781}
& \underline{0.797}
& 0.945
& 0.945
& 0.945 \\

GPT-OSS-SFT
& 0.961
& \underline{0.898}
& \underline{0.844}
& 0.961
& 0.961
& 0.961 \\

GPT-OSS-RL
& 0.938
& \underline{0.734}
& \underline{0.680}
& \underline{0.930}
& 0.938
& 0.945 \\

\hline
Average
& 0.847
& \underline{0.738}
& \underline{0.730}
& \underline{0.843}
& \underline{0.847}
& 0.850 \\
\hline
\end{tabular}
}

\vspace{1mm}

\textbf{(c) JamC-QA}

\resizebox{0.82\textwidth}{!}{
\begin{tabular}{p{0.20\linewidth}|*{6}{>{\centering\arraybackslash}p{0.12\linewidth}}}
\hline
Model & Orig. & Trans. & Repl. & Homo. & IME & FullW. \\
\hline
\hline

Ministral-8B
& 0.352
& 0.367
& \underline{0.328}
& 0.352
& 0.352
& 0.352 \\

Swallow-8B
& 0.461
& \underline{0.430}
& \underline{0.383}
& \underline{0.453}
& 0.469
& 0.469 \\

ELYZA-8B
& 0.398
& \underline{0.367}
& \underline{0.359}
& 0.398
& 0.398
& 0.398 \\

llm-jp-13B
& 0.508
& 0.516
& \underline{0.453}
& \underline{0.469}
& 0.508
& 0.508 \\

llm-jp-7.2B
& 0.508
& \underline{0.445}
& \underline{0.391}
& 0.523
& 0.508
& 0.508 \\

Sarashina-3B
& 0.578
& \underline{0.531}
& \underline{0.516}
& 0.578
& 0.586
& 0.586 \\

Sarashina-1B
& 0.313
& 0.344
& 0.313
& \underline{0.305}
& 0.313
& 0.320 \\

Sarashina-0.5B
& 0.273
& 0.289
& 0.281
& 0.289
& 0.273
& 0.289 \\

GPT-OSS-20B
& 0.297
& 0.336
& \underline{0.250}
& 0.297
& \underline{0.289}
& 0.305 \\

GPT-OSS-SFT
& 0.531
& \underline{0.453}
& \underline{0.492}
& \underline{0.508}
& 0.547
& \underline{0.500} \\

GPT-OSS-RL
& 0.383
& 0.414
& \underline{0.344}
& 0.391
& 0.445
& 0.445 \\

\hline
Average
& 0.418
& \underline{0.408}
& \underline{0.374}
& 0.416
& 0.426
& 0.426 \\
\hline
\end{tabular}
}

\end{table}

\label{app:task_results}

\begin{figure}[tb]
\includegraphics[width=\linewidth]{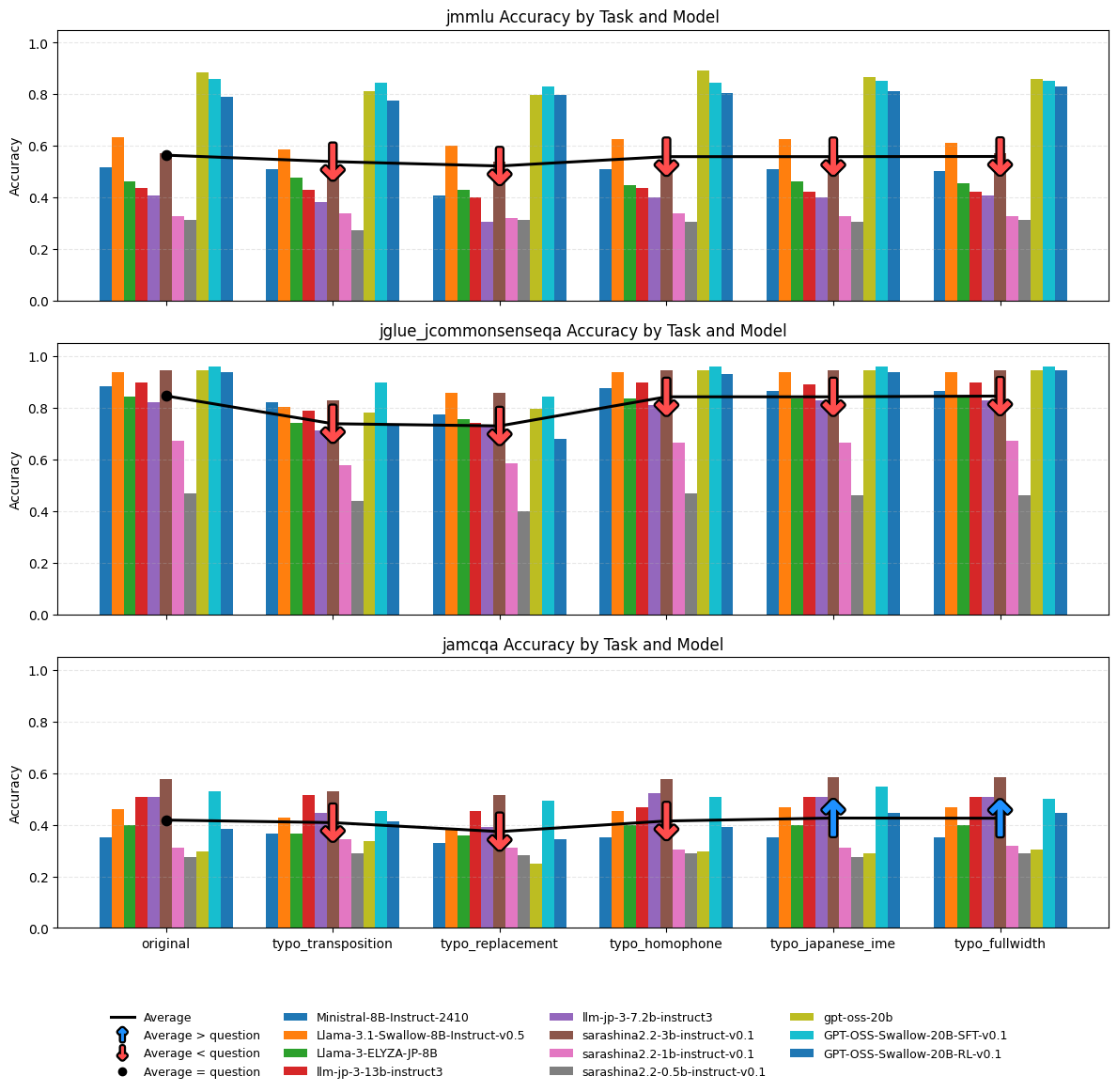}
\caption{Accuracy of each model under different typo conditions for JMMLU, JCommonsenseQA, and JamC-QA. The black line denotes the average accuracy across all evaluated models, and arrows indicate changes from the original condition.}\label{fig:results}
\end{figure}

\bibliographystyle{splncs04}
\bibliography{references}

\end{document}